# Rethinking and formalising the state across languages: a unified computational learning theory account

**Mohamed El Idrissi**
mohamed.elidrissi@inalco.fr
INALCO, France

**Abstract**

The linguistic notion of state has traditionally been restricted to the construct (annexation) state of Afroasiatic languages and treated as a language-specific morphosyntactic phenomenon. This article argues instead that the state is a systemic, context-dependent morphosyntactic mechanism that selects grammatical templates across synthetic languages. Within the Template-Based Modular Cognitive framework, taking Riffian as its primary empirical basis, the proposed theory provides a unified explanation for diverse nominal marking patterns traditionally analysed independently and is formalised as a symbolic computational model in which the state is represented by a set-valued function over grammatical templates. A learning algorithm based on finite-set operations acquires and predicts state-dependent grammatical configurations. Beyond nominal morphology, the framework has broader implications for theories of nominal structure and lexical cognition, in particular offering a unified analysis of determiner–noun structure. These results suggest that the state constitutes one instance of a broader class of syntactically conditioned dependencies that also includes agreement and grammatical case.

**Keywords:** computational linguistics, morphosyntactic theory, semantic theory, countability, definiteness, grammatical case, construct state, symbolic learning

## 1 Introduction

Word-forms of lexemes vary in different ways, particularly in nouns. The best-known type of variation is inflection, but this represents only a small part of the variation attested across languages. Variation can also occur at the inter-word level (see Table 1 and 3.2.10) and the morphosyntactic-shift level, including overt anti-morphosyntactic shift (El Idrissi 2024). A given variation may affect only a single paradigm of the lexeme (e.g., gender, number); conversely, a single paradigm — especially definiteness and countability — may be affected by all of these variations. A further characteristic is that grammatical variation can be triggered by the syntactic context of the lexeme, as with agreement and grammatical case.

**Table 1: Comparison of nominal marking patterns in Riffian and English illustrating inter-word variation**

| | Riffian | | | English | |
|---|---|---|---|---|---|
| | Marker | Radical | | Marker | Radical |
| CN | *a* | *zˤru* | stone | the | woman |
| | N/A | *fad* | thirst | N/A | water |
| PN | N/A | *mammat* | Mammat | N/A | Agatha |
| | *a* | *zlaf* | Azlaf | the | Rockies |

Those differences are in connection with another phenomenon involving the same morphemes. Thus, depending on several syntactic positions (e.g., with the coordinating conjunction *and*), the bare form is preferred instead of the article or the state. This pattern is discovered in various languages, especially in synthetic languages. Subordinated syntactically to these conditions, a marker switching occurs between either two markers or a marker and an empty site on the surface. In (1), the example *by boat* expressed in different languages exemplifies the pattern in question:

(1) In English: by **Ø** boat (< **the** boat)
In Riffian: gi ð-**Ø**-ʁerabut (< ð-**a**-ʁerabut)
In French: en **Ø** bateau (< **le** bateau)
In German: per **Ø** Schiff (< **dem** Schiff)
In Tagalog: sakay **ng** bangka (< **ang** bangka)
In Greek: με **Ø** πλοίο (< **το** πλοίο)
In Hebrew: b ***Ø*** sira (< ***ha*** *sira*)
In Wolof: *ak gaal* ***Ø*** (< *gaal* ***gi***)

We argue that, despite their surface differences, these phenomena are characterised by similar irregular marking patterns, suggesting a shared underlying structure across languages. We therefore propose a theoretical model to account for these phenomena. These morphological shifts thus arise at two levels: inter-word (see Table 1) and in situ (see (1)). Some languages have both (e.g., English,

Riffian), while others exhibit only the in-situ variation (e.g., Greek). Various explanations have been proposed in Semitic, Romance, and Germanic linguistics to account for these morphological contrasts, especially for the inter-word variation. A traditional view attributes them to the supposed semantic properties of nouns. For instance, proper nouns (PNs) are often considered inherently definite, rendering the definite article redundant. More broadly, such semantics-driven approaches attempt to explain marking patterns, but they fail to capture the full range of irregularities, notably the in-situ alternation. They are typically language-specific and limited to particular noun classes (e.g., kinship terms, uncountable nouns, PNs, or countable nouns), aiming more to justify individually observed forms than to unify them. This calls for a reassessment of such phenomena. Notably, the definite article is rarely treated as an integral morphological element (El Idrissi 2024).

Our goal is to identify the linguistic characteristics of nouns and to detail the morphosyntactic differences among nominal lexemes. It also addresses broader theoretical questions about the nature of nouns and their morphological and semantic properties. In this light, this study proposes an alternative framework to account for the in-situ, inter-word, and inter-language morphosyntactic variations. While addressing specific thematic areas, the article primarily analyses the state, defining it as a dependency-based relationship, which constitutes a pivotal element in understanding the morphological variations discussed above. To demonstrate the theory's broader implications, we use the Riffian language as our empirical foundation, offering a novel perspective on the state that has remained largely unexplored in existing literature. The primary contribution of this paper is a formal computational theory of the linguistic state.

Throughout this article, we adopt the Template-Based Modular Cognitive (TBMC) framework, previously introduced by El Idrissi (2024, 2026), as the formal basis for modelling the proposed theory of the state. Its architecture is designed as a learning framework in which the mathematical model is parameterised by discovering the relevant parameters from observed data. Owing to the specific properties of the state phenomenon, additional learners have been developed to model state shifts by exploiting inter-word variation. These learners demonstrate how the proposed computational learning framework can be applied to state markers. Through this formalisation, several consequences follow from this theory. We show that determiners can be analysed as noun phrases, thereby contributing to a unified syntactic analysis.

Hence, the concept of state is reconsidered and disentangled from its markers; on that basis, we propose a unified, formalist framework for morphological noun marking. We have implemented decisive notions to characterise different observed phenomena that, if taken as they appear in traditional grammars, could obscure the cross-linguistic analysis. The core of this theoretical study is about markedness and the newly developed concept of state. We have shown how nouns are marked with respect to countability and definiteness markers by taking several languages into account. We compared the differences in marking among nouns, notably for linguistic varieties that exhibit a morphological switching involving the singulative/the definite and collective meaning.

This study proceeds as follows: Section 2 reviews the relevant morphological literature and introduces the theoretical framework. Section 3 first provides a formal definition of the state and its computational properties and then analyses cross-linguistic data to establish the relationship between markedness, state, and morphological templates within a unified framework. Finally, Section 4 concludes the study by summarising the theoretical and empirical findings.

## 2 Background and theoretical foundation

The following sections provide a critical review of the literature, focusing on the theoretical foundations of nominal marking and the structural specificities of the Riffian system. Furthermore, we analyse the traditional treatments of the state in Riffian and theoretical linguistics to identify the conceptual gaps that our proposal seeks to address. The review also includes a discussion of the principal frameworks in learning theory.

### 2.1 Riffian nominal morphology: Theoretical background

Riffian is a synthetic, largely fusional language. Consequently, it utilises both derivation and inflection to generate new lexemes and their corresponding word-forms. Thus, each morpheme can refer to different grammatical meanings. The general canonical form of a noun is composed of these elements:

| noun = countability + radical[1] + gender + number |
|---|

As illustrated in Table 2, the masculine singular typically functions as the unmarked form[2][3]. Morphological markers can also appear as circumfixes; for example, the feminine singular marker is realised at both the left (*ð-*) and right (-t) edges of the radical.

---

[1] Terminologically, a radical must be distinguished from a root. Defined in structural opposition to inflectional morphemes, a radical can comprise a root morpheme, a scheme morpheme, and any derivational morphemes. Note that alternative terms are frequently employed depending on the theoretical school.

[2] List of abbreviations employed in this study: SG = Singular; PL = Plural; F = Feminine; M = Masculine; V = Verb; VP = Verb Phrase; N = Noun; NP = Noun Phrase; COL = Collective; PART = Partitive; GEN = Genitive; SING = Singulative; DAT = Dative; INSTR = Instrumental; DET = Determiner; PREP = Preposition; NEG = Negative.

[3] All examples have been transcribed in IPA notation.

**Table 2: Number and gender marking on nouns**

| | |
|---|---|
| *a-kidaɾ* | horse.SG.M |
| *ða-kidaɾt* | mare.SG.F |
| *i-kidaɾen* | horse.PL.M |
| *ði-kidaɾin* | mare.PL.F |

Countability is not a universal descriptive category across languages. Many authors have debated the nature of these morphemes (El Hankari 2014), which traditional Riffian literature calls either case or state. We avoid the term case because we consider this grammatical class unrelated to case marking (see 3.2.11). Instead, we draw a clear distinction between countability and state, treating them as two separate levels of linguistic description.

First, the countability markers (see Table 3), always prefixed in Riffian, comprise two morphemes, namely the collective (e.g., ∅-*fus* 'hand'), that is a null marker (symbolised by ∅), and the singulative morpheme (e.g., *a-ɾgaz* 'man'). This element contributes to the inter-word variations seen in Table 1. Crucially, the feminine marker inflects within the countability paradigm, onto which it is morphologically affixed. Within this paradigm, the singulative marker exhibits overt marking, whereas the collective marker remains unformed (see 3.2.7) and uninflected. Together, these markers constitute a structural unit independent of the nominal base. Such a configuration closely mirrors the behaviour of gender-marking definite articles in Romance languages. The singulative, which has a phonological form, also varies in number and gender. Either they are assigned to words on the basis of their countable features, or their affixation to the radical is governed by arbitrary rules (e.g., lexical changes).

**Table 3: Countability paradigm in Riffian**

| | Singular | | Plural | |
|---|---|---|---|---|
| | Feminine | Masculine | Feminine | Masculine |
| Collective | t-∅ | (w/j)-∅ | NA | NA |
| Singulative | t-a | a | t-i | i |

As for the state, this article introduces a new perspective that will be expanded upon in Section 3. To ground this theory, we first offer a descriptive analysis of the state's properties within the Riffian language. Thus, the state acts on the countability markers by inflecting them; in other words, the singulative marker is replaced by the collective marker. One may observe that the countable nouns can be both singulative and collective, which seems contradictory at first glance. Such a phenomenon is syntactically driven, much like agreement or grammatical case, and directly accounts for the observed in-situ variations. Consequently, conditioned by the syntactic environment (Lafkioui 2007, pp. 113–115), nouns in Riffian alternate between two primary morphological states, traditionally designated as the free state (FS) and the annexation state (AS). The scenarios where countable nouns have a collective meaning/form correspond to situations where SUBSETHOOD (see 3.2.8) is borne by another morpheme such as with the partitive or pseudo-partitive construction. For instance, in Riffian, numerals, and other words expressing quantity, are in fact nouns, more precisely pro-forms acting as unit words, employed almost exclusively in the contexts cited above. Another case, when the singulativity is relegated to favour collectivity, is for instance with prepositions. In such constructions, the collective is required:

(2) ðɾaθa n je-∅-qzinen (FS: i-qzinen)
three of COL-dogs
'Three dogs'

(3) s-bɾa ðe-∅-wːaɾθ (FS: ða-wːaɾθ)
with-NEG COL-door
'Without door'

Furthermore, in the annexation state, a semi-vowel (*w* or *y*) appears exclusively in masculine nouns. Our thesis is that *w-*/*j-* is just a phonological artefact. Thus, this morphophonological phenomenon is triggered solely when these specific syntactic constructions are satisfied and affects words that carry either the collective or singulative marker, as demonstrated in (2).

(4) Collective marker:

*a.* Free State:

∅-fus a-fusi
COL-hand SING-right
'the right hand'

*b.* Annexation state:

a-ʁɾum n **w**-∅-fus
SING-bread GEN COL-hand
'homemade bread'

(5) Singulative marker:

*a.* Free State:

a-ɾgaz n zik

SING-man GEN early

'The man of yesteryear'

*b.* Annexation state:

a-wal n **w**-∅-ɾgaz

SING-word GEN COL-man

'man's word'

In the annexation state, a central vowel may surface in the position of the singulative marker. However, this vowel lacks phonemic contrast, as this epenthesis is driven entirely by syllabification constraints. This may yield the vowelisation of the semi-vowels (see Table 4) if complex phonological patterns converge. As established, the phenomenon is not restricted to morphophonology; it functions at the morphosyntactic interface.

**Table 4: State shift in nouns**

| Free state | Annexation state |
|---|---|
| *a-qzin* | *u-∅-qzin* ⇐ /we-qzin/ |
| *ð-a-qzint* | *ðe-∅-qzint* |
| *i-qzinen* | *je-∅-qzinen* ⇐ /we-qzinen/ |
| *ð-i-qzinin* | *ðe-∅-qzinin* |

In the next sections, we explore the implications of this new concept by examining how the state manifests across different languages and structural levels. We first review how the state has been treated across different fields of linguistics.

## 2.2 The state in traditional and theoretical linguistics

In linguistics, the bare term state is not generally used on its own; rather, it was borrowed from Semitic studies, where the established terminology distinguishes between the construct state and the absolute state. This terminology was later adopted in the study of other language families (e.g., Andersen 2002; Smirnova 1982), although it was not necessarily applied to the same linguistic phenomena (Chaker 1995).

Traditional grammars of Semitic languages describe the construct state as a phenomenon restricted to the genitive construction, in which the head noun lacks a definite marker. In all other contexts, the absolute state is used. In Berber studies, including Riffian linguistics, the concept of state was retained, but the terminology was modified. The construct state was renamed the annexed, annexation, or bound state, whereas the absolute state became known as the free state.

The differences, however, are not merely terminological. In Berber languages, the annexation state occurs not only in genitive constructions but also in a range of other syntactic contexts, particularly those associated with grammatical functions. Its occurrence in contexts involving grammatical functions (Ouhalla 1996) has led some scholars to reinterpret the phenomenon as a form of case marking rather than as a state distinction (see 2.1). A similar tension between the notions of case and state has been observed in the description of other language families. It is also worth noting that some authors use the term construct case, thereby conflating the two concepts (Mous 1993; König 2008).

Within the framework of generative grammar, the structural derivation of the construct state was initially formulated through N-to-D movement (Ritter 1991), under which the head noun raises from N to D to acquire the definiteness value of its genitive complement. This analysis became theoretically problematic with the advent of the Minimalist Program (Chomsky 1995), as subsequent minimalist developments imposed stricter constraints on head movement. In particular, the Extension Condition (Chomsky 1999) rendered classical head adjunction increasingly difficult to reconcile with the architecture of narrow syntax, thereby casting doubt on analyses that relied on N-to-D movement. This theoretical shift created a significant challenge for explaining the obligatory adjacency and head-initial order characteristic of construct state nominals.

More broadly, minimalist syntax responded to these architectural concerns by replacing many derivational operations based on head movement with a system centred on phases and Internal Merge. Within this framework, Chomsky (2000, 2001, 2004) proposed that phase heads may bear an OCC (i.e., OCCurrence; also referred to as Edge) feature, which triggers Internal Merge by attracting constituents to the edge of the phase. Because the Phase Impenetrability Condition (PIC) restricts direct access to material within completed phases (i.e., Strict Cyclicity Principle), syntactic movement must proceed cyclically through successive phase edges. The OCC feature therefore provides a general mechanism for successive-cyclic movement, accounting for diverse displacement phenomena, including object movement and Quantifier Raising, within a unified phase-based architecture.

For analyses of the construct state, this minimalist reconceptualisation substantially reshaped the theoretical landscape. As the status of classical N-to-D movement became increasingly controversial, researchers sought alternative explanations that either replaced head movement with phrasal movement or derived the construct state outside narrow syntax. Aligning with the minimalist preference for Internal Merge, Shlonsky (2004) proposed a phrasal movement analysis in which the nominal phrase, rather than the noun itself, moves to a higher specifier position. By contrast, Siloni (2001) argued that the construct state is derived through postsyntactic operations at the Phonetic Form (PF) interface, thereby avoiding the need for a narrow-syntactic derivation. Finally, Borer (1999) advanced a lexical approach in which the construct state reflects the inherent definiteness properties of nominal heads, treating construct morphology as a lexical property rather than the output of syntactic movement. This proposal was developed specifically for Semitic languages and has not been widely generalised.

This study follows neither of these traditions. Instead, it seeks to enlarge the concept of state, moving beyond its restriction to Afroasiatic or African languages and treating it as a universal linguistic feature found in diverse, unrelated synthetic languages. Section 3 presents our exact definition of the state and its theoretical scope. The primary framework for this article is the Template-Based Modular Cognitive model (El Idrissi 2024, 2026). This model uses learning theory, specifically computational algebraic learning, to capture the linguistic behaviours of synthetic languages.

### 2.3 Related work in learning theory: Symbolic and Sub-symbolic

Learning theories fall into two broad paradigms, differing in how knowledge is represented and in the nature of the rules or parameters that the learner infers from data. Despite differing terminologies, every learning system decomposes into the same two functional components: a learner, a fixed procedure whose goal is to infer underlying rules or parameters from data, and a predictor[4], built from those inferred rules, whose goal is to map new inputs to outputs. What differs between paradigms is the form these rules or parameters take: continuous numerical parameters in the sub-symbolic paradigm, or discrete symbolic objects in the symbolic paradigm. Despite the emergence of hybrid approaches, the symbolic/sub-symbolic distinction remains a central organising framework for machine learning and artificial intelligence.

In the sub-symbolic paradigm, the learner infers continuous numerical parameters — such as vectors, matrices, or real-valued functions — by optimising an objective function over a continuous search space. The resulting predictor then applies these parameters to new inputs. This paradigm encompasses statistical learning theory (Vapnik 1998), neural networks, kernel methods, and deep learning. Theoretically, these methods are grounded in probability theory, measure theory, functional analysis, convex optimisation, and information theory. Furthermore, the predictor's generalisation is characterised through empirical risk minimisation, regularisation, Vapnik–Chervonenkis dimension, and statistical consistency (Vapnik 1998; Mohri et al. 2018).

[4] Depending on the framework in Artificial Intelligence, it may be called differently, for instance: a classifier, a grammar, a hypothesis, a model, etc.

In the symbolic paradigm, the learner instead infers discrete symbolic objects — such as logical formulas, algebraic structures, grammars, automata, graphs, or finite sets — that satisfy logical, algebraic, or combinatorial constraints. The resulting predictor then applies these symbolic objects to classify or generate outputs for new inputs. This approach draws heavily on mathematical logic, universal algebra, automata theory, graph theory, and computational complexity. Representative areas include inductive logic programming (Muggleton & De Raedt 1994; Nienhuys-Cheng & De Wolf 1997), grammatical inference (de la Higuera 2010), exact learning (Angluin 1987), algorithmic learning theory (Gold 1967), and program synthesis (Gulwani et al. 2017).

Beyond this representational split, the paradigms diverge most notably in three further respects. First, the predictor's generalisation is conceived differently: the sub-symbolic predictor offers probabilistic guarantees on unseen data, while the symbolic predictor targets exact identification, consistency, or recoverability of the symbolic objects it infers (Gold 1967; Angluin 1987). Second, interpretability diverges sharply. Symbolic objects are explicit and directly inspectable, whereas the numerical parameters inferred by a sub-symbolic learner encode knowledge implicitly and resist direct interpretation. Finally, the two treat uncertainty differently: the sub-symbolic predictor models stochastic variation probabilistically, while the classical symbolic predictor assumes deterministic reasoning, though statistical relational learning and probabilistic logic programming now bridge this gap (De Raedt et al. 2016).

The framework proposed in this work belongs to the symbolic tradition. More specifically, it constitutes an algebraic symbolic learning approach in which the learner, termed the gradient condition, infers a set parameter, a symbolic object represented as a finite set endowed with the operations of a Boolean algebra. The predictor then applies this parameter to derive morphological outputs. Learning is formulated as the inference of this unknown set parameter through algebraic identities involving set operations, particularly the symmetric difference. This operation forms an abelian group and, together with intersection, constitutes a Boolean ring (Halmos 1963; Stone 1936). Rather than optimising continuous numerical parameters, the proposed framework learns a discrete algebraic object satisfying a collection of equations derived from morphological evidence. This perspective distinguishes the proposed framework from both statistical learning theory and logic-based symbolic learning by placing algebraic structures at the core of the learning process.

# 3 Cross-linguistic and theoretical perspective

In Table 3, we have displayed the countability category that has different forms and senses, plus or minus countable, according to the state. Hence, the state change consists of alternation between the collective and singulative marking, to put it in another way AS and FS. Some nouns will always have one form for both states, as is the case with some uncountable nouns (U) in Riffian, and others will alternate between one form or another, like countable nominals (C). The state must be distinguished from word formation processes involving the prefixation of the singulative marker on the uncountable nouns when creating new countable lexemes. In one case, we deal with a contrast between two different word forms, whereas in the second, we have two separate lexemes. Of course, that is specific to Riffian; another language may apply the same to other kinds of nouns, if we are interested in the prospect of extending the concept of state to other languages. We consider the state and countability two different concepts, since we assume that the state can apply to other grammatical features or lexical categories (e.g., verb). This also goes for countability and number; each has its own morphemes and the markers of countability can be either singular or plural. Besides, the collective marker can be used with the singular or the plural, which means they are indeed different paradigms. A noun with an uncountable meaning ignores inflection in both number and countability because both need to be associated with a noun that can be discretised to do so. Their neutralisation in the same semantic surroundings is just coincidental; in other circumstances, they may inflect separately. Thus, countability is a grammatical class of nominals, like the classifier, the case and so forth, that deals with the discretisation/subsets spectrum when it comes to quantity (see 3.2.8). As to the state, it is similar to the case marking or the agreement, in that the marking on a noun will change according to another morpheme/lexeme where a dependency is engaged between both entities.

## 3.1 Formalising the state

### 3.1.1 Preamble

We think that the state, as an unrecognised/overlooked linguistic fact at the interface of morphosyntax/syntax and morphosemantics — and sometimes also morphophonology —, may be discovered in other languages with other grammatical classes (e.g., Overt/covert: ±X/∓Ø or Overt/overt: ±X/∓Y; X and Y are the forms of grammatical markers). We have in mind, for example, the augment (Katamba 2006), found in Bantu languages, which also appears to be under the sway of the state (e.g., In Zulu, AS of *u-muntu* ‘person’ in vocative case: ***Ø****-muntu* ‘O person’). Thereby, in addition to this extension to other grammatical paradigms, we propose that the morphosyntactic or syntactic relationships activating marker switching may be very limited or vast (as in Riffian). The context does not need to be alike in all languages to recognise the state; there are, however, several similar typological patterns present in different unrelated languages (see (1)). For instance, the definite article, a primary focus of this study, serves as a clear example of a marker that shares the same morphosyntactic properties as the singulative in Riffian (see (2) - (5)). We will not develop these contrasts at length here; we wish only to illustrate our definition of the state by taking a fresh look at other languages. We give some simple examples where the marking surfaces whether it be on the head or the dependent by neutralising the use of the definite marker:

(6) in French:

- with the instrumental preposition *en* (e.g., AS: *en* ***Ø*** *bois* 'in wood' / FS: ***le*** *bois* 'the wood'; AS: *en* ***Ø*** *avion* 'by plane' / FS: ***l'****avion* 'the plane')
- with the genitive preposition *de* (e.g., AS: *revue de* ***Ø*** *littérature* 'literature review' / FS: *revue de* ***la*** *littérature* 'review of the literature')

(7) in English:

- with the instrumental preposition *by* (e.g., AS: by **Ø** plane' / FS: **the** plane)
- with the genitive marker *s* or genitive preposition *of* (e.g., AS: the man's **Ø** hat / FS: **the** hat of the man; AS: a slice of **Ø** cheese / FS: eat **the** cheese; AS: the hardness of **Ø** wood / FS: the hardness of **the** wood)

### 3.1.2 The set-theoretic definition of the state and learning model

This study seeks to unify the analysis of these features. The constructions illustrated above are variously labelled in the literature as the zero article (Jespersen 2007) or bare nouns (de Swart et al. 2007), and they remain the subject of extensive theoretical debate. However, the pattern highlighted by our study has a larger scale that overlaps partly with these debates. The absence of the definite article is more a secondary matter if we take a step back to observe different morphological realisations as a whole. The state can also be stationary (i.e., non-template shift) or non-stationary (i.e., template shift). The stationary case is observed at the FS as well as the AS. The stationary state in AS occurs because of the inter-word variation when some items have the same template as the FS.

#### 3.1.2.1 The mathematical formalism of the state

The state is defined as a mechanism that relates a set of grammatical markers to a set of words and syntactic contexts. In Definition 2, the basic assumption of this concept is represented by a function.

**Definition 2:** Let us consider this definition of the state

- Let $L$ be the set of the languages[5],
- Let $M$ be the set of meanings
- Let $W$ be the set of morphemes
- Let $C$ be the set of syntactic contexts
- Let $T$ be the set of grammatical templates such that $T \subset \mathcal{P}(M)$
- Let $state: C \times T \rightarrow T$
- $\exists\, l \in L, \exists\, c \in C, \exists\, t_i \in T$ such that $state(c, t_i) = t_j$ where $t_i$ represents the unmarked template and $t_j$ the AS or FS's template, $t_j \in T$.

The previous piecewise-defined function has several branches $state_n(c, t_i) = t_j$, $n \in \mathbb{N}$, and is integrated structurally to the morphosyntactic transfer function $h(w)$ (El Idrissi 2024, 2026). Hence, the output of $state_n(c, t_i)$ is paired with the input $w \in W$ to form an item-template pair: $(w, t_j)$. As part of the Template-Based Modular Cognitive model, a learning phase and prediction phase are required. Thus, $t_j$ is predicted by learning first the operand set $p_n \subset M$ via the learner, the gradient condition (El Idrissi 2024, 2026). Thus, the operand set is computed beforehand from the gradient condition and determines the branch-specific transformation applied to the template. When $p_n$ is determined, this parameter is used to compute $t_j$ by applying the predictor, namely the $\mathrm{state_n}(c,\ t_i)$, which is endowed with the symmetric difference operation: $t_j = t_i \,\Delta\, p_n$. In the context of this study, two branch functions $state_1$ and $state_2$ operate, specific to languages, to inflect the free state and annexation state, as exemplified in Example 1.

**Example 1:** The following examples from Riffian (excluding the morphophonological semi-vowel phenomenon, see 2.1) and French illustrate the prediction of nouns with different unmarked templates and states.

- <u>In Riffian:</u>

  A) AS and FS different:

  - AS of *a-rgaz* in the genitive case (spell-out form: *Ø-rgaz*, e.g., *a-wal n Ø-rgaz* 'man's word') with $p_{rif,1} = \{-COL, +SING, +COL, -SING\}$:

$$state_{rif,1}(GEN, \{\mathrm{N, +SG, -PL, +M, -F, -COL, +SING}\}) = \{N, +SG, -PL, +M, -F, +COL, -SING\}$$

[5] The set $W, C, T$ and the function *state* are genuinely language-indexed. From a cross-linguistic perspective, a family of pairs $(W_l, T_l)$ and a family of functions $state_l$: $C_l \times T_l \rightarrow T_l$ for $l \in L$, would better reflect this. However, since we work systematically with a small, fixed sample of languages, we suppress the index and simplify the notation accordingly.

- FS of *a-rgaz* in the object case (spell-out form: *a-rgaz*, e.g., *zˤriʁ a-rgaz* 'I saw the man') with $p_{rif,2} = \{\}$:

$$state_{rif,2}(OBJ, \{\text{N}, +\text{SG}, -\text{PL}, +\text{M}, -\text{F}, -\text{COL}, +\text{SING}\}) = \{N, +SG, -PL, +M, -F, -COL, +SING\}$$

B) AS and FS identical:

- AS of *Ø-fus* in the subject case (spell-out form: *Ø-fus*, e.g., *iqːen Ø-fus* 'the hand is closed') with $p_{rif,2} = \{\}$:

$$state_{rif,2}(SUBJ, \{\text{N}, +\text{SG}, -\text{PL}, +\text{M}, -\text{F}, +\text{COL}, -\text{SING}\}) = \{N, +SG, -PL, +M, -F, +COL, -SING\}$$

- FS of *Ø-fus* in vocative case (spell-out form: *Ø-fus*, e.g., *aː Ø-fus* 'O hand') with $p_{rif,2} = \{\}$:

$$state_{rif,2}(VOC, \{N, +SG, -PL, +M, -F, +COL, -SING\}) = \{N, +SG, -PL, +M, -F, +COL, -SING\}$$

- In French:

C) AS and FS different:

- AS of *la-voiture* in vocative case (spell-out form: *Ø-voiture*, e.g., *ô voiture* 'O car') with $p_{fr,1} = \{-COL, +DEF, +COL, -DEF\}$:

$$state_{fr,1}(VOC, \{\text{N}, +\text{SG}, -\text{PL}, -\text{M}, +\text{F}, -\text{COL}, +\text{DEF}\}) = \{N, +SG, -PL, -M, +F, +COL, -DEF\}$$

- FS of *la-voiture* in object case (spell-out form: *la-voiture*, e.g., *Je vois la voiture* 'I see the car') with $p_{fr,2} = \{\}$:

$$state_{fr,2}(OBJ, \{\text{N}, +\text{SG}, -\text{PL}, -\text{M}, +\text{F}, -\text{COL}, +\text{DEF}\}) = \{N, +SG, -PL, -M, +F, -COL, +DEF\}$$

D) AS and FS identical:

- AS of *Ø-Paul* in vocative case (spell-out form: *Ø-Paul*, e.g., *ô Paul* 'O Paul') with $p_{fr,2} = \{\}$:

$$state_{fr,2}(VOC, \{\text{N}, +\text{SG}, -\text{PL}, +\text{M}, -\text{F}, +\text{COL}, -\text{DEF}\}) = \{N, +SG, -PL, +M, -F, +COL, -DEF\}$$

- FS of *Ø-Paul* in subject case (spell-out form: *Ø-Paul*, e.g., *Paul court* 'Paul runs') with $p_{fr,2} = \{\}$:

$$state_{fr,2}(SUBJ, \{\text{N}, +\text{SG}, -\text{PL}, +\text{M}, -\text{F}, +\text{COL}, -\text{DEF}\}) = \{N, +SG, -PL, +M, -F, +COL, -DEF\}$$

### 3.1.2.2 The selection of the branch functions

To complete the computation of the branch functions during the prediction phase, two conditions are required to select the branches to apply (i.e., either $state_1$ or $state_2$). A second condition is essential because, in some languages, like French and Riffian, some nouns have the same template at the AS and FS (see B) and D) in Example 1). Such a particularity is caused by the inter-word variation (see also Table 1). Hence, this selection cannot be based only on the syntactic context. The applicability condition is defined as the conjunction $Cond_1 \wedge Cond_2$. The parameters governing both conditions are learned during the training phase, and the conditions are subsequently evaluated during the prediction phase to determine the appropriate branch function:

- $Cond_1$: c ∈ $C_q$ such that $c \in C,\ C_q \in \{C_{FS}, C_{AS}\}, where\ C_{FS} \cap C_{AS} = \emptyset$ and $C_{FS}, C_{AS} \subset C$. $C_{FS}$ is the set of the free state's contexts and $C_{AS}$ is the set of the annexation state's contexts. Only the $C_q$ for which $Cond_1$ evaluates to true is retained; this is the decision rule given to the learner of $Cond_1$.

- $Cond_2$: $p_u \setminus t_i = d_v$ such that $u, v \in \mathbb{N}$, $t_i \in T$ is the unmarked template, $p_u \subset M$ is the learned operand set defined as $p_u = \underset{p \in P}{\operatorname{argmax}}\ \#p$, where $P \subset \mathcal{P}(M)$ denotes the family of candidate operand sets of the state, # is the cardinality. As for $d_v \subset M$, its values are learned by computing $d_{learning}(p_u, t_i)$ = $p_u \setminus t_i$.
  For example, in Riffian, $p_1 = \underset{p \in P}{\operatorname{argmax}}\ \#p$ where $p_1 = \{+COL, -SING, -\mathrm{COL}, +\mathrm{SING}\}$ and $d_v$ is either equal to $d_1 = \{+COL, -SING\}$ or $d_2 = \{-COL, +SING\}$.

During learning, the values of $p_u$ and the corresponding difference set $d_v$ are learned and stored as model parameters. During prediction, only the input template $t_i$ and the syntactic context $c$ vary, while the learned parameters $p_u$ and $d_v$ are used to determine the appropriate branch function. In Example 2, the prediction phase for the Riffian language is demonstrated.

**Example 2:** The following examples from Riffian illustrate how the morphological system selects the $state_n$ functions. We reuse the examples employed in Example 1. As one can see below in E) and F), the correct operand set $p_n$ is used with $t_i$.

- In Riffian:

  E)

  - AS of *a-ɾgaz* in the genitive case (spell-out form: *Ø-ɾgaz*, e.g., *a-wal n Ø-ɾgaz* 'man's word') applies $state_{rif,1}$ only under the following conditions:

$$if\ GEN \in C_{AS}\ and\ \{+COL, -SING\} = \{+COL, -SING, -\text{COL}, +\text{SING}\} \setminus \{\text{N}, +\text{SG}, -\text{PL}, +\text{M}, -\text{F}, -\text{COL}, +\text{SING}\}\ then$$

$$\{N, +SG, -PL, +M, -F, -COL, +SING\}\ \Delta\ \{-COL, +SING, +COL, -SING\} = \{N, +SG, -PL, +M, -F, +COL, -SING\}$$

- FS of *a-ɾgaz* in the object case (spell-out form: *a-ɾgaz*, e.g., *zˤriʁ a-ɾgaz* 'I saw the man') applies $state_{rif,2}$ only under the following conditions:

$$if\ OBJ \in C_{FS}\ and\ \{+COL, -SING\} = \{+COL, -SING, -\text{COL}, +\text{SING}\} \setminus \{\text{N}, +\text{SG}, -\text{PL}, +\text{M}, -\text{F}, -\text{COL}, +\text{SING}\}\ then$$

$$\{N, +SG, -PL, +M, -F, -COL, +SING\}\ \Delta\ \{\} = \{N, +SG, -PL, +M, -F, -COL, +SING\}$$

F)

- AS of *Ø-fus* in the subject case (spell-out form: *Ø-fus*, e.g., *iqːen Ø-fus* 'the hand is closed') applies $state_{rif,2}$ only under the following conditions:

$$if\ SUBJ \in C_{AS}\ and\ \{-COL, +SING\} = \{+COL, -SING, -\text{COL}, +\text{SING}\} \setminus \{\text{N}, +\text{SG}, -\text{PL}, +\text{M}, -\text{F}, +\text{COL}, -\text{SING}\}\ then$$

$$\{N, +SG, -PL, +M, -F, +COL, -SING\}\ \Delta\ \{\} = \{N, +SG, -PL, +M, -F, +COL, -SING\}$$

- FS of *Ø-fus* in vocative case (spell-out form: *Ø-fus*, e.g., *aː Ø-fus* 'O hand') applies $state_{rif,2}$ only under the following conditions:

$$if\ VOC \in C_{FS}\ and\ \{-COL, +SING\} = \{+COL, -SING, -\text{COL}, +\text{SING}\} \setminus \{\text{N}, +\text{SG}, -\text{PL}, +\text{M}, -\text{F}, +\text{COL}, -\text{SING}\}\ then$$

$$\{N, +SG, -PL, +M, -F, +COL, -SING\}\ \Delta\ \{\} = \{N, +SG, -PL, +M, -F, +COL, -SING\}$$

#### 3.1.2.3 The computational implementation of the learning and validation phase

The proposed function is intended to apply to any synthetic language. The overall architecture is language-independent, whereas the sets W, C, T, and the branch functions are language-specific. Algorithm 1 implements the proposed learner and can be adapted to specific languages by defining C, T, and D (i.e., the dataset of the observed free states ($t_x$) and annexation states ($t_y$) of items). The generic learner consists of four main parameter-learning functions (see the Supplementary Material for a more detailed version of Algorithm 1 – Appendix A and a code implementation – Appendix B): *EvaluateCond₁*, *LearnCandidateOperandSet* and *DLearning*, which determine the parameters of the conditions used to select the appropriate branch, and *GradientCondition*, which determines the parameters of the selected branch required to compute the AS template. Given at least one training pair with each of $state_1$ and $state_2$ realised (see Examples 1 and 2), the learner recovers the parameters exactly.

**Algorithm 1 (Condensed): Learner of the state's parameters**

---

**Require:** $c \in C$; $t_i, t_j \in T$; $C_{FS}, C_{AS} \subset C$ ($C_{FS} \cap C_{AS} = \emptyset$); $D = \{(t_x, t_y)\}$
**Ensure:** Retained subset $C_q$, gradient $p_n$, condition $d_v$

1 **Initialize:** $C_q, p_u, p_n, d_v \leftarrow \emptyset$
2 **for** $C_{cand} \in \{C_{FS}, C_{AS}\}$ **do**
3     **if** $c \in C_{cand}$ **then** — ▷ *Evaluate Cond₁*
4         $C_q \leftarrow C_{cand}$
5         $P \leftarrow \{t_y \Delta t_x \mid (t_x, t_y) \in D\}$ — ▷ *Build candidate space*
6         $p_u \leftarrow \text{argmax}_{p \in P} \#p$ — ▷ *Maximal operand set*
7         $p_n \leftarrow t_j \Delta t_i$ — ▷ *Instance gradient*
8         $d_v \leftarrow p_u \setminus t_i$ — ▷ *Extract $Cond_2$ via $p_u$*
10     **end if**
11 **end for**

12 **if** $C_q \neq \emptyset$ **then**
13     **return** $(C_q, p_n, p_u, d_v)$
14 **else**
15     **return** Drop c
16 **end if**

---

To evaluate the proposed learner, an independent validation dataset $D_{test} = (t_i, \hat{t}_j)$ was constructed from the Riffian language. Unlike the training dataset D, which is used to extract the retained parameter sets, $D_{test}$ evaluates two complementary stages of the learner independently:

- Rule-selection validation, which assesses whether the dual-condition execution mechanism correctly selects or rejects retained rules.

- Target-state validation, which assesses whether the selected rule reconstructs the expected target state.

This separation is important because successful rule selection does not necessarily imply that the observed target state is structurally admissible. Consequently, rule execution and state prediction are evaluated independently.

**Table 5: The validation dataset consists of six representative cases**

| Case | Description | Rule-selection objective | Prediction objective |
|---|---|---|---|
| 1 | Observed exact match | Execute retained rule | Correctly reconstruct observed target |
| 2 | Unobserved equivalent context ($C_{FS}$) | Generalise to unseen context within ($C_{FS}$) | Correctly reconstruct observed target |
| 3 | Unobserved equivalent context ($C_{AS}$) | Generalise to unseen context within ($C_{AS}$) | Correctly reconstruct observed target |
| 4 | Unobserved invalid context | Reject rule execution | No prediction |
| 5 | Unobserved invalid baseline state | Reject rule execution | No prediction |
| 6 | Cross-contaminated state | Select the structurally matching rule | Reject incompatible target state |

Rule selection is governed exclusively by the execution conditions of $Cond_1 \wedge Cond_2$. A retained rule is executed only when both conditions are satisfied. The first condition verifies contextual-subset membership, whereas the second verifies compatibility between the baseline state and the learned structural transformation. Once a rule has been selected, the selected branch predicts the target state (i.e., AS) by applying the retained set parameter $p_n$: $\hat{t}_j = t_i \Delta p_n$. The predicted state $\hat{t}_j$ is then compared with the observed target state $t_j$. The validation therefore measures two complementary criteria (see Supplementary Materials for the code implementation – Appendix B). Our validation was performed on the Riffian language; the learner established four rules including the transition case and the non-transition cases. Table 6 details the accuracy of the learner for different cases.

The first evaluation concerns the correctness of the execution mechanism (see Table 5). Cases 1–3 verify that the learner successfully retrieves the appropriate retained rule for both previously observed and previously unseen contexts belonging to the corresponding contextual subsets. Cases 4 and 5 verify that execution is correctly blocked whenever either the membership condition or the applicability condition is violated. Case 6 further verifies that the learner retrieves the rule associated with the structural characteristics of the input, although its primary purpose is to evaluate the accuracy of the predicted target state.

The second evaluation assesses the accuracy of the predicted target state after rule execution (see Table 5). For Cases 1–3, the predicted state exactly matches the observed target state, confirming that the retained parameters correctly reconstruct both observed transitions and previously unseen but structurally equivalent contexts. In contrast, Case 6 intentionally provides a target state that is incompatible with the rule selected by the execution mechanism. Although the learner correctly selects the applicable rule according to the conditions, the predicted state differs from the supplied target state, demonstrating that the learner does not reproduce structurally inconsistent target representations (see Table 6).

**Table 6: Summary of the validation results**

| Case | $Cond_1$ | $Cond_2$ | Rule selected | $\hat{t}_j = t_j$ | Interpretation |
|---|---|---|---|---|---|
| 1 | True | True | Executed | Yes | Correct rule selection and correct prediction |
| 2 | True | True | Executed | Yes | Generalisation within $C_{FS}$ - Stationary case-FS |
| 3 | True | True | Executed | Yes | Generalisation within $C_{AS}$ - Stationary case-AS |
| 4 | False | — | Rejected | — | Correct rejection (invalid context) |
| 5 | True | False | Rejected | — | Correct rejection (invalid baseline template) |
| 6 | True | True | Executed | No | Correct rule selection; incompatible target state detected |

### 3.1.3 The determiner paradigm as a noun phrase

Other situations in both languages could have been added (e.g., in French: *tout* 'all'; in English: *each*). These latter morphemes are generally not considered as triggers of the AS, but rather as they are in mutual exclusivity with other determiners to which they would belong as well. We would like to propose instead that the existence of this paradigm has to be contested, and the presence of these morphemes causes the absence of the definite article (i.e., triggers the AS) as a preposition could do (e.g., *by car*). In other words, the definite article must not belong to the class of determiners and must instead be considered as part of the morphological domain as theorised. Thus, we claim that the surface form [J ⊕ K], as in French *chaque étudiant* 'each student', where ⊕ is the syntactic operation, J corresponds to specific determiners (e.g., *chaque* 'each') — excluding the definite article — and K to nouns (e.g., *étudiant* 'student'), is treated as the output of an underlying representation containing a definite element: Surface [J ⊕ K] is derived from underlying J ⊕ {the + K} → J ⊕ {∅ + K}, because {the + K} and J are considered the unmarked forms.

In French and English, there is no article on J; we assume that the collective marker is privileged. On K, by contrast, the definite article or the collective marker can be used depending on the morpheme J. But in other languages, the definite marker or another equivalent marker can be affixed both on J and K such as in Basque or Maori (e.g., In Maori: *le tausaga taitasi* {the year each} 'each year'; In Basque: *ikasle guzti-ak* {students all-the.PL} 'all the students'). As observed from a cross-linguistic perspective, the underlying structure of [J ⊕ K] is not restricted to the bare forms; both J and K can have all the grammatical features, including definiteness and countability. Therefore, from a topological perspective, we assume a more generic underlying {Ø/the + J} ⊕ {Ø/the + K} that may surface [{DEF + J} ⊕ {DEF + K}] as in Hebrew (quantitative adjective, for instance, *rav* 'many'), [{DEF+ J} ⊕ K] as in Basque, [J ⊕ {the + K}] as in Maori, [J ⊕ K] as in English. A specific language can show all the mentioned syntactic surfaces; they are not mutually exclusive. In English, both [J ⊕ {the + K}] (e.g., all the people) and [J ⊕ K] (e.g., every man) exist. This is entirely governed by the state triggering the FS or AS depending on syntactic contexts (see 3.1.2).

If accepted, we need to presume that between J and K a functional relationship exists. This is crucial; we will come back to this point later. Furthermore, the previous illustrations involve morphemes being in the orbit of NPs, but this is not mandatory. A marker licensing the state on a NP may be affixed to the VP (e.g., In Wolof (Diouf 2003 p.100) with the applicative construction, especially the instrumental applicative *-e*: *woto la dem-e {car CM leave-INSTR}* 'He leaves by car'/ **woto bi la dem-e*). Hence, this also mirrors the behaviour of some determiners that can be positioned inside the verb phrase in some languages (e.g., In English, *The students have all left*), a phenomenon known as Quantifier Float (Sportiche 1988). Such a parallelism reinforces our position in analysing adpositions and determiners as syntactic contexts triggering the state or being subject to the state.

The corollary of this proposal is that it challenges some aspects of the DP or QP analysis (Etxeberria and Giannakidou 2019), since all of these constructions can instead be analysed as NPs. The generic underlying {Ø/the + J} ⊕ {Ø/the + K}, which can surface in different ways through the state, as seen earlier, must be analysed as an NP-NP construction rather than QP–DP or QP–NP, where the internal structure of NP is the item-template pair. This can be abstracted all the way up from surface to syntactic structure. The abstraction proceeds in two stages: (i) modelling of the surface string, in which the linear form is decomposed into two units and then abstracted to item–template pairs; and (ii) mapping from the underlying representation to the abstracted surface, in which $state_2$ or $state_1$ computes each item's state-marked template from its unmarked template and syntactic context, while the non-state-marked item–template pairs are separately realised as the syntactic representation NP ⊕ NP.

i. Modelling of the surface:

i.a. The surface string is decomposed into two units with their respective state-marked forms, combined by the syntactic operator ⊕:

1) all the students ⟹ {Ø + all} ⊕ {the + students}

2) each student ⟹ {Ø + each} ⊕ {Ø + student}

i.b. Each unit is abstracted to an item–template pair, as theorised in the TBMC model:

1) {∅ + all} ⊕ {the + students} ⟹ (all, {N, +COL, −DEF, …}) ⊕ (students, {N, −COL, +DEF, …})

2) {∅ + each} ⊕ {∅ + student} ⟹ (each, {N, +COL, −DEF, …}) ⊕ (student, {N, +COL, −DEF, …})

ii. From abstracted surface to underlying structure representation:

ii.a. The state takes as input each item's unmarked template and its syntactic context (HEAD or GEN), returning the corresponding state-marked template:

1) (all, {N, +COL, −DEF, …}) ⊕ (students, {N, −COL, +DEF, …}) ⟹ (all, state$_2$(HEAD, {N, +COL, −DEF, …})) ⊕ (students, state$_2$(GEN, {N, −COL, +DEF, …}))

2) (each, {N, +COL, −DEF, …}) ⊕ (student, {N, +COL, −DEF, …}) ⟹ (each, state$_2$(HEAD, {N, +COL, −DEF, …})) ⊕ (student, state$_1$(GEN, {N, −COL, +DEF, …}))

ii.b. The non-state-marked item–template pair is realised as the syntactic representation:

1) (all, state$_2$(HEAD, {N, +COL, −DEF, …})) ⊕ (students, state$_2$(GEN, {N, −COL, +DEF, …})) ⟹ NP ⊕ NP

2) (each, state$_2$(HEAD, {N, +COL, −DEF, …})) ⊕ (student, state$_1$(GEN, {N, −COL, +DEF, …})) ⟹ NP ⊕ NP

This considerably simplifies the syntactic analysis and calls into question the developments proposed around the OCC feature, which are unnecessary in our theory because no Move operation / Internal Merge is involved. Furthermore, from this perspective, inter-word variation does not appear to be restricted to a subset of synthetic languages. Languages previously thought to be exempt from this phenomenon, such as Greek, no longer constitute exceptions. In Greek, some determiners (e.g., *amfóteroi* 'both' and *ólos* 'all') never co-occur with the definite article, whereas all nouns, including proper nouns, bear it as their unmarked form (e.g., *ho Giórgos* 'Georges', *to vivlío* 'the book'). Thereby, this inter-word is not uniquely observable between PNs and CNs, but it can also be discovered among other lexemes. Like many other languages, Greek also exhibits several state constructions involving determiners: [{the + J} ⊕ K] (e.g., *to káthe prósopo* 'each person'), [J ⊕ {the + K}] (e.g., *óli oi fitités* 'all the students'), and [J ⊕ K] (e.g., *pollés gynaíkes* 'many women').

The determiner itself as a linguistic category is questionable; in our view, it is no more different than the adjectives or nouns from a morphosyntactic standpoint, since they can agree in number and gender or be marked by the definite article (e.g., In Basque, *gizon bakoitz-a* {man each-the} 'each man'). Nouns can also be exempted from any word-form like with some determiners, as typically observed in Riffian with loanwords (see El Idrissi 2024). A strong position would be to reject this category outright. For now, this question is left open, and is outside the scope of our study.

Let us limit our demonstration to these fragments that are enough to explain our purpose. The few instances given imply definiteness, as in Semitic, as seen before. We are aware that French and English studies do not see these relevant patterns as connected and similar to what happens in Semitic and even less in Riffian. Even in Semitic studies, the notion of the construct state (i.e., AS) is only contemplated with the genitive construction, whereas it has been noticed that this label could be used in other environments where the definite article is absent; for example, in Arabic, after the words *bidu:n* or *bila*: 'without'. Let us mention that Arabic is an example of a language where the AS is restricted to a few cases, contrary to English or Riffian.

Throughout this article, we do not attempt an exhaustive treatment of each language, because, by focusing on individual cases, we may lose sight of the whole by addressing the sum of the individual parts. Instead, we propose a general framework for the state. Nevertheless, each language has its own particularities, which may obscure or complicate the mechanism formalised in Definition 2. These differences are too numerous to be examined within the scope of a single article.

For these reasons, let us provide a general framework setting the limits of the state more precisely in the next section. But for the sake of clarity, without loss in generality, we will mention several languages to corroborate our statements. As the state is articulated and closely intertwined with markedness and formedness, this also provides an opportunity to further clarify these notions.

### 3.2 Theory of Marking: Markedness, Formedness and State

Our framework claims that different lexical features are encoded natively in different modular templates. Thus, one of them, countability, as a grammatical category, is observed in various languages, often associated erroneously with number. With regard to marking, the singulative seems to be the one that is the most cited in the world's languages (Greenberg 2013) — because the collective is often unformed (see below). To understand how Riffian is situated with respect to other languages, we should distinguish marking in terms of markedness and formedness. Those two concepts, although related, are strictly different and necessary to the description of linguistic facts. It is needed because the passage from the unmarked form to the marked form may be realised by morphological pruning (e.g., disfixation, clipping or an unformed marker) or zero derivation. Terminologically, we prefer to speak of a (un)marked form instead of a (un)marked morpheme, because markedness may interact with an expletive morpheme.

In our view, the bare noun is not systematically the unmarked one. Some (Haspelmath 2006) consider them analogous terms, since, in their view (Jakobson 1990 p.158), the marked form is always the one having an additional formed morpheme or a suprasegmental/non-concatenative feature added to a morpheme regarded as the unmarked form. It is true that the terms encoded/coded, marked and formed are used interchangeably in the literature. Nevertheless, we would like to make a clear distinction between those concepts. In our conception of morphology, when we talk about markedness, we rely on a process of word and meaning formation that has a beginning and an ending. By taking into account these processes, we are able to distinguish the morphological features. Thus, the unmarked form denotes the input material and the marked one the output one. Typical processes used between the starting and the final step are, for example, conversion, inflection, semantic shift, morphological derivation or the state.

### 3.2.1 Inflectional versus derivational markedness

Inflectional and derivational markedness should also be considered in the analysis of this phenomenon. When taking into consideration this derivational aspect, it is crucial to acknowledge that the subclasses of a part of speech are autonomous from one another. Thus, the uncountable and the countable categories must be distinguished at the cognitive level, because a word and meaning formation process can occur between them. Moreover, a markedness relationship holds between two words and is not established by considering all the words in a syntactic paradigm (inflection) or across two paradigms (derivation). In this latter view of markedness, the most frequent form, or the bare noun, would be treated as the unmarked form (Greenberg 1980). By contrast, in our view, markedness is defined in an item-based pairwise interaction. Let us take the gender marking as an example. Thus, if a language has animate referents that, inside a syntactic paradigm, can distinguish between female and male and encodes the feminine on the basis of the masculine by adding a morpheme, it does not mean that any feminine word belonging to this paradigm has to be then analysed as the marked form. In illustration, let us contemplate the following examples: *ða-funast* 'cow.F' is marked, because it is built upon *a-funas* 'bullock.M', and, in contrast, the word *ð-ixsi* 'ewe.F' is unmarked although it is feminine; its masculine counterpart uses the suppletive form *ikeṛṛi* 'sheep.M'. Also, the feminine uncountable noun *t-sam:erθ* is unmarked; the marked form is the masculine countable noun *sam:er*. In addition, the verb *sum:er* is unmarked in comparison to the marked form *t-sam:erθ* that derives from it. Finally, with respect to the grammatical template, our notion of markedness operates both at the level of the template as a whole and at the level of its individual grammatical features.

### 3.2.2 Modular templates and markedness

This series of examples could make us think that we are proposing an atomistic approach to markedness; in reality, we conceive that markedness is inherent to a paradigm. Each of these paradigms includes words that share the same word and meaning formation process and some semantic values, especially animacy in Riffian. In addition, each one has a template of predefined-unmarked forms (see 3.1). The subcategories mentioned before do not necessarily fit the distinction between uncountable and countable nouns or even between common and proper nouns; they are at a lower level than these previous categories. We do not analyse them as being unrelated, that is why we use the term subcategory in order to indicate explicitly that they can be put together in a vaster category. These modular subsets have a cognitive basis, that is why they had been termed COGNITIVE SET.

The words of these cognitive sets have grammatical templates. We already cited some of them. For example, the unmarked marking of NA, as an autonomous category, contains the masculine gender, the singulative countability and the singular number. This is not generic to all subcategories; another one may have other unmarked forms (e.g., Deverbal U are feminine in gender and collective in countability). Though a NA does not certainly refer to an animate material object in place in the world's objects, but rather to an event without actors and aspects (in contrast, verbs encode events with their actors and aspects), those markers are still present. That being said, they are not expletive forms; the meaning conveyed by these unformed morphemes is formally expressed and perceptible in several ways, notably the agreement. The dissymmetry is not situated at the interface between semantics and morphology, but rather between semantics and the non-cognitive world. Needless to say, the world's reality should not be the starting point of linguistic analysis. What we should retain from these remarks is that any subcategory necessarily has its own morphological template regardless of the (perceived) trustworthiness vis-à-vis reality. For this reason, we do not endorse the semantics-driven marking approach. The only situation where there is a disruption in markedness is when the word and meaning formation process used was conversion or when a semantic shift occurred. From a cross-linguistic standpoint, depending on each language, the number of existing modular nominal subcategories can be restrained or extended.

### 3.2.3 Intrinsic versus extrinsic markers

Therewith, it is also required to make allowances for, when speaking about markedness, two kinds of markers that are affixed to the radical, namely intrinsic inflectional markers/paradigms and extrinsic ones. For the case of the Riffian noun, the first one, the only one relevant to markedness (see hereinafter), encompasses gender, number and countability (see 2.1), while the second kind refers more or less to what is known as the determiners. It is usual to consider the definite article as belonging to the latter, but it seems more appropriate to attach it to the former. In Riffian, definiteness is not encoded, but in other languages, this category is primordial in matters of markedness as it will be unveiled thereafter (see further below this point). Moreover, the intrinsic inflectional markers shall also be divided into inner intrinsic markers and outer ones. This division is pertinent for describing the fossilisation of the countability and number markers after conversion (see 3.2.5), which does not occur with gender (i.e., gender shift) regarded here as an outer intrinsic marker. The concept of intrinsic inflectional marker can be compared with that of noun class (Dixon 1968), but their conceptual positions are not totally aligned. The latter can be criticised on two grounds. First of all, it has a restricted conception of what might be regarded as such. Only certain inflectional paradigms found in some limited languages are included in this linguistic category. This makes this notion a peculiar phenomenon, while, in our view, we consider that this is not particular to some languages, but it is a universal linguistic fact. What may be language specific is the grammatical paradigms that can be regarded either as an intrinsic inflectional paradigm or not, but they should have at least one. Because of this, we include number, definiteness and countability, which is not usually viewed in this way. Moreover, we think that the intrinsic inflectional markers are not merely restricted to the noun; verbs or other parts of speech also have their own intrinsic markers. Secondly, our concept is intimately linked to markedness and state. Therefore, inside those paradigms, there is necessarily an unmarked form and, potentially, one or several marked forms that can be triggered by a given set of syntactic contexts.

### 3.2.4 Borrowed intrinsic markers and the scope of the cloning process

That leads us to also suggest an identical reasoning for the morphological reanalysis of the Arabic and Romance definite articles that are borrowed along with the radical in Riffian. It is generally advanced that this article has become an extension of the radical. But this does not justify why this morpheme, in particular, is retained in comparison to others. Our proposition is that we could analyse conversion and borrowing as processes entailing the same cognitive operation(s), because, in both cases, grammatical morphemes are viewed as expletive forms. Then, it would be more appropriate to recommend that what enables the borrowing of this article is its (re)analysis as an inner intrinsic marker in the same way as the countability markers. Furthermore, when the loanword has no affixed definite marker as with some nouns of address, we suggest that the unformed collective marker is used (e.g., *Ø-meskin* (↰ in Arabic: *al-miskin*) or *Ø-puβri* (↰ in Spanish: *pobre*) 'poor one'). In light of this, we argue, for nominal words, that the minimal scope of borrowing or conversion covers the intrinsic inflectional markers present in the lexical matrix.

### 3.2.5 In derivational markedness, intrinsic markers also behave as inflectional markers

Accordingly, there is then nothing strange about a word or sub-paradigm having the two labels (e.g., with a deverbative noun, the noun is marked; with a denominal verb, the noun is unmarked). In fact, the marked form really makes sense only in inflectional markedness, especially when derivation occurs between two different parts of speech. In derivational one, the marked form is also always the unmarked form of the inflectional markedness. The use of the term marked for this case is made in opposition to the starting subcategory having served to form the subcategory in question. This is particularly true for Riffian where countability plays a role in both inflection and derivation. The unmarked form is not necessarily the same in both morphosyntactic processes. When the noun is uncountable, the collective marker is the unmarked form and the derived countable noun affixed with the singulative marker is the marked one. In contrast, in the inflectional environment, the singulative marker is the one that is unmarked and the collective is the marked form (see Example 1 in Section 3.1).

Whether it be the gender or the singulative marker, they are not derivational markers, as stated before, but inflected forms. As the singulative marker has the possibility to vary in number and gender, that excludes the hypothesis of a derivational marker. This latter, in terms of morphology, is part of the radical; it is not affected by these kinds of grammatical features. Their inflectional nature is eminently obvious when the noun is referring to animate referents, especially to gender. Therefore, it seems questionable to assume they are derivational markers. In reality, as viewed before, the derivation between the uncountable and the countable nouns is done by conversion, in other words, by zero derivation. The addition of a morpheme in a derivational context should not be interpreted automatically as the manifestation of a derivational marker. Its presence is the consequence of the unmarked form shift, especially from an unformed to a formed marker, resulting from the subcategory change after derivation (i.e., a template shift corresponding to the transition to another template having different unmarked forms, see also 3.2.8).

### 3.2.6 Free and conditional variation between markers used with the state

As researchers infer the cognitive representations of languages from the realisation surface, it will be no simple matter to determine the one that is unmarked or marked. Given that, conversion, affixation and pruning can participate in word formation, and this also happens on the surface, unformed analytic encoding[6] and fusional encoding are difficult to dissociate if no morphophonological cue is left. In order to fathom this problem, other linguistic features than morphological ones could be useful for this task. Moreover, in some languages, the state shift does not seem to follow a rigid rule. In Riffian, a countable noun preceded by a preposition, for instance, will mandatorily use the collective marker. In French, by contrast, the collective meaning can be in free variation with the definite article in some cases (e.g., *sans la voiture* 'without the car' / *sans voiture* 'without cars', but **en la voiture* / *en voiture* 'by car'). Thereby, this free variation may be constrained and only possible within a defined range of morphosyntactic/syntactic environments and/or modular subcategories.

Thus, the free, in-situ and inter-word variations must be differentiated when analysing a linguistic variety. Hence, free variations of those markers (e.g., singulative/collective;

[6] Let us recall that unformed markers can be regular members of an inflectional paradigm.

definite/collective) should be analysed as instances of classical morphological inflection, such as the distinction between singular and plural, and are unrelated to the state. From a computational standpoint, they are not triggered by the state function, but the inflectional system. In addition, some words, although being able to be employed in the two word-forms without constraint, may be more frequently used in one word-form (i.e., collective marker) than the other (i.e., definite marker). We can mention to illustrate, for example, in English: *school, home*, *bed*, etc (see Soja 1994); in Italian: *scuola* 'school', *casa* 'home', *letto* 'bed', etc. However, there is then no reason to assume that these nouns have a different unmarked form. Thus, we reject the hypothesis that those nouns have similar patterns to PN or U as postulated by Longobardi (2001). These nouns in question are not intrinsically different from other countable nouns. Besides, even though the speaker is on certain occasions free to choose one marker or the other, the AS is only activated when uttered in syntactic contexts, which makes the notion of state a crucial concept for linguistic description.

### 3.2.7 Template shift, markedness and formedness

This free variation may also be just an appearance because we are dealing with different lexemes that have the same radical form. In this case, we have several subcategories with different unmarked forms. This may not seem obvious to the casual eye, since when a category change occurs, this process is accompanied generally by an ambivalent markedness shift. The latter, in English or other languages, may be systematically related to countability that is in the orbit of the state. As countability may be involved in different types of variation (inflection/free variation, state, inter-word, etc), this diversity of manifestations can lead researchers to overlook its underlying role. This confusion is accentuated in English and other languages alike because no grammatical gender shift occurs after conversion (see El Idrissi 2026). In Riffian, by contrast, gender, neutral with regard to the state, will play this role by shifting when the item undergoes conversion. This enables us to rule out any possible confusion in Riffian, while, in other languages, it is more difficult to perceive this subcategory change on the surface.

By bearing all of that in mind, it becomes easier to understand why, for instance, some words can come with the definite article, and in other cases, they may not. To illustrate our purpose, let us take an example. In English, the word *nature* (according to the Bloomfield classification (Bloomfield 1933, p.205), this word is part of class II.B.2) is generally considered to have one form and several meanings (i.e., fusional encoding) and depending on its meaning and its state (i.e., conditioned by sets of syntactic contexts), the definite article may be used or disregarded. But the analysis that we performed in this class is different. Actually, there would be two different lexemes belonging to two separate subcategories. We could distinguish the uncountable noun, when it refers to the physical world (e.g., *I love nature*), from the countable one (Bloomfield's class II.A) meaning the quality/the property of something (e.g., *I love the nature of X*). In our view, the latter derives from the former by conversion (U → C) and their marking is different regarding countability and definiteness (($nature_1$, {N, - DEF, + COL, + M, - F, + SG, - PL}) → ($nature_2$, {N, + DEF, - COL, + M, - F, + SG, - PL})).

This word formation process affects the unmarked form of U, which changes when U becomes C. The unmarked form of *nature*, as a U, is collective in countability and non-definite in definiteness. In contrast, after the markedness shift, its countability is collective, its definiteness is definite and its marked form in some contexts (e.g., with the genitive construction *-s*) is collective and non-definite. Moreover, such a distinction permits us to prevent the confusion of the last paradigm with another U paradigm mentioned by Bloomfield. These words are classed by Bloomfield in class II.B.1 represented by the word *milk.* They are distinguishable from class II.B.2 by different properties. When both are U, their relation to definiteness is different: *half of nature* / **half of the nature* vs *half of the milk.* We think that their distinctiveness comes from their underlying encoding. Thus, *nature*, when U, has fused different meanings under one form, whereas, with *milk,* the meanings are dissociated and bound to different morphemes (see El Idrissi 2024). The previously proposed analysis is just an outline of what could be done in other languages distinct from Riffian. Then, the probable subcategories existing in English are more varied. Therefore, for each language, it is necessary to establish a formation model of U (see El Idrissi 2024) or any lexeme in order to determine the unmarked forms of each subcategory. Most crucially, this permits us to avoid computing a morphological prediction paradox.

As one can observe, words of class II.B.2 have an unformed morpheme expressing the collective meaning. However, although this pattern is widespread across the world's languages, it is not universal. Some languages instead exhibit formed collective marking. In traditional grammatical descriptions, however, these markers are generally misidentified, often being analysed as genitive or case markers (see also Section 3.2.11 for a discussion of formed collective markers and case marking). For example, in Estonian, the markers traditionally described as genitive markers occur not only in genitive constructions but also in other syntactic contexts, such as after a preposition (e.g., *tänu sõbra abile* 'with the help of a friend') or as the marker of a direct object (e.g., *ma nägin raamatu* 'I saw a book'). We propose instead that these morphemes be analysed as collective markers. The existence of formed collective marking in some languages, in turn, supports the view that unformed collective marking is a genuine cognitive reality.

### 3.2.8 The Countability-Definiteness nexus

Pending further study of each of these languages, we would like nevertheless to tentatively suggest that the mentioned observations could be generalised to all the synthetic languages subject, of course, to adaptation to each linguistic variety with regard to the template, markedness and state. In Riffian, the state involves solely the countability paradigm, namely the collective and the singulative, while, in other languages, such as French or English, it interacts with the countability and definiteness paradigm. This paradigm may share the same semantic class, but these markers are not necessarily part of the same inflectional class. More concretely, we can assume that the partitive and the collective belong to the same semantic class, but when spelled out, they can inflect together.

In theory, this could be possible, but it has not so far been documented. In Estonian, it is not clear if both markers are mutually exclusive, since with some words they can be used together, for example: *raamat-u* {book-GEN} / *raamat-u-t* {book-GEN-PART}. This interpretation is valid only if these markers are indeed analysed as partitive morphemes, a point that remains controversial. The interaction between definiteness and countability, however, is less disputed. Therefore, neither is mutually exclusive of the other, in principle. For example, in Welsh, the singulative and the definite article can be used together: *yr-ader-yn* the-bird-SING 'the bird'. This implies that their nature regarding markedness may be different and then, they may be considered, independently of one another, either intrinsic (e.g., the definite article *l-* in French) or extrinsic markers that are not relevant for markedness (e.g., the preposition/partitive marker *de-* in French).

Nonetheless, inside these two inflectional classes, two analytic markers, the collective and the definite, are never used together in the languages having both paradigms. Their opposition is particularly clear in several languages. We may suppose that, given that both express the wholeness of their respective sets, the collective and the definite meaning are equivalent in this sense and may represent the counterpart of one another. We theorise that countability and definiteness, from a semantic point of view, are related to the notion of SUBSETTABILITY. It means that, if subsettable, a set representing a referent is perceived as divisible into subsets; this particularity is marked by the countability markers. In contrast, if non-subsettable, the definite marker is used; in this respect, the set is not internally divisible by the same referent, but the partition is introduced externally by another lexeme (e.g., portion). The two lexemes stand in a subsethood relation, whereby one is interpreted as denoting a subset associated with the other.

The distinction between *un exemple de maison* 'an example of house' and *un exemple de la maison* 'an example of the house' further illustrates the notion of subsettability. In the former, *maison* denotes a type whose members constitute a subsettable set; *un exemple* selects one instance from the class of houses. In the latter, *la maison* refers to a specific referent that is construed as a whole rather than as a member of a class. Here, *un exemple* is interpreted as an illustration, copy or representation of that referent, not as one of several instances. Consequently, the opposition cannot be reduced to mere partitionability (as in *une portion d'eau* vs *une portion de l'eau*), since neither construction involves the physical division of a referent. Instead, subsettability captures a more general semantic distinction between expressions that license the extraction of a member from a conceptual set and those that identify a unique, non-subsettable referent. Singulativity is the atomic realisation of subsettability, in which the selected subset consists of exactly one individuated element, whereas partition constructions instantiate non-atomic realisations.

### 3.2.9 Undeterminedness and markedness

Let us point out that, when referring to unformed, this includes the two kinds of encoding cited in the previous paragraphs, but as to formed, only the formed analytic encoding is considered for a grammatical feature. This formed marker could also be a fusional marker encoding different grammatical features, but it is not bound to the radical. Furthermore, we do not support the idea that any inflectional morpheme must necessarily be described with regard to markedness. This notion is only relevant when an inflectional intrinsic marker appears or disappears subsequent to an induced change or when the markers are always involved in derivational processes (e.g., from C to U) targeting the morpho-phonological layer. If the marker answers only to a need for finer semantic distinctions (e.g., demonstrative), then it must be viewed as an extrinsic marker. For example, trying to find which one is the marked form among spatial deictic markers (e.g., proximal and distal, respectively *–a* & *–in* in Riffian) is futile, because they do not satisfy the aforementioned requirements at least in Riffian. In this case, the unmarked template of the inflectional paradigm can be considered undetermined. Nevertheless, their undeterminedness is not perpetually cemented; when used with another part of speech or subcategory, the templates can be classified as marked or unmarked. Hence, these markers should be analysed instead as intrinsic markers. For example, in Riffian, with the pronouns, the spatial deictic paradigm is in this instance part of the intrinsic markers.

### 3.2.10 Inter-word and in-situ variations between PN and CN

Moreover, as the collective and definite meaning can fulfil the same intent, either can be used with PN. Thus, in Modern Greek or Tagalog (distinct from the one used with CN), the definite article is preferred instead of the collective as the unmarked form. Although, in these languages, the unmarked form of this subcategory is the definite marker, those PNs, like other nouns, have a marked form, induced by the state, that is the collective marker (e.g., in Modern Greek: *Ποια Ρώμη* 'which Rome' / * *Ποια η-Ρώμη* 'which the Rome', see also (1)). This in-situ variation is not observable in languages, such as Riffian or English, where the unmarked form of a subcategory, namely proper or uncountable nouns, is identical to the marked form in AS. The cross-linguistic capacity to affix definite articles to PNs supports our model's premise, that in specific morphological templates, the collective value of a proper noun is the primary grammatical feature being encoded. Therefore, PN's intrinsic markers are not limited to gender and number; they also encompass countability. A shared perception of the physical world might suggest a universal linguistic encoding; however, grammatical categories are not mere reflections of empirical reality. Each lexeme is associated with intrinsic markers that mandate the selection of an unmarked form within its paradigm. This template acts as the organising principle of morphology, independent of the perceptible world.

Thus, in keeping with the inter-word variations, as a marking shift between U and C can occur, this is also possible with PN (e.g., In Riffian: *ʒeħ:a* 'PN' → *a-ʒeħ:a* 'mischievous person'). In Riffian, the markedness of PN and CN is different; the singulative marker is affixed to the latter, while the former has the same unmarked form as U, that is to say, the collective meaning. While comparable patterns are attested cross-linguistically, the specific markedness configurations differ. In Germanic and Romance languages, the definite article can co-occur with PNs, as demonstrated by examples like '*The United Kingdom*', while most PNs are determinerless. A related phenomenon occurs with deonymised PNs, such as French la bougie ('candle') stemming from the article-less toponym Bougie. Under these structural circumstances involving proper names, the marker functions as an expletive determiner. Nevertheless, in the AS, the collective meaning is used as with regular nouns (e.g., *Victoria's United Kingdom*).

### 3.2.11 Confusion between case marking and state

Let us turn to another crucial point related to the state. As said before, the morphosyntactic/syntactic contexts triggering the AS may be different depending on languages, but there is a specific context of a particular interest present in several languages on which we would like to say a few words. In some languages, when a lexeme is in a core argument position (object or subject), the collective marker is preferred to the unmarked form. English is one of them (Carlson 1977) and this is also what happens in Riffian. The collective meaning/marker is used in such a wide range of situations that it is inconceivable that this marking be only related to a syntactic function. This remark can be deployed to other languages where the same analysis has been proposed for some markers, while their link to definiteness and/or countability is undeniable. For example, Tagalog uses the formed morpheme *ng*, referred to by some as a genitive marker (Kroeger 1991), in various syntactic positions: argument position, after a preposition, in a genitive construction, after a numeral, etc.

In the same way, the state and countability are often confused with the case; that happens regularly when the markers are formed, unlike Riffian, French or English where the collective meaning is unformed. Thus, in different Finnic languages, several markers, which are again syntactically conditioned, are used to express a meaning related to countability. This paradigm includes the genitive marker, which could be reinterpreted as a collective marker, and the partitive marker. They may be in free variation, for instance, in object position (e.g., In Estonian;

Genitive/collective: *ta sõi leiva* 'He ate bread' / Partitive: *ta sõi leiba* 'He ate (some) bread'), which clearly indicates that their main function is not to mark the grammatical function of nouns. In these previous languages, the AS, which is associated with the collective meaning, is favoured, but for the same syntactic functions, in other languages, the FS prevails instead as in Albanian where the definite marker is used (e.g., non-definite and collective: *zog* 'bird'; definite: *zog-u* 'the bird'; definite and accusative case: *zog-u-n* 'the bird').

A reanalysis of these languages is then welcome on the basis of structural elements of our theory: markedness/formedness, fusional/analytic encoding, states in nexus with syntactic conditions, word and meaning formation processes, and distributed subcategories shaped by a template. The case marking and the state should be clearly distinguished; a case marker, which could be unformed as well (Legate 2008), is only intended to be related to a syntactic function. However, those syntactic-based markings are not in contrast to each other; they could be used together as in Riffian (e.g., *i-u-qzin* {DAT-COL-dog} 'to the/a dog'). In fact, as the state is bound to the intrinsic markers, it is incumbent on the unmarked or marked form to mark the noun; it is a peremptory template. Then, it is expected to discover both markers affixed next to each other when a case marker is employed. Moreover, if the case paradigm is also a mandatory element of a template (i.e., an intrinsic marker), then we can predict that the unmarked case marker, especially if it is formed, may be found in various syntactic positions, although its function does not require it to be present.

### 3.2.12 Syntactic dependency

The state is a syntactically conditioned inflectional phenomenon, as evidenced by the cross-linguistic data presented throughout this study. This does not imply that every alternation involving definiteness or countability should be analysed as an instance of the state, as discussed in Section 3.2.6. Rather, the state is triggered only in specific syntactic contexts. Although these contexts are not identical across all synthetic languages, they exhibit substantial overlap, particularly in core argument positions (e.g., subject and object), adpositional constructions, head–modifier constructions, and related syntactic configurations.

It is worth mentioning that the state does not apply exclusively to either the independent or the dependent lexeme; rather, it may affect both, as shown in Section 3.1. This accounts for the cross-linguistic observation that the same phenomenon (i.e., the omission of a morpheme, such as the definite article, in a given syntactic context) may target either the head, as in the Semitic construct state, or the modifier, as in Berber.

The state function applies to all nominal lexemes. Whenever a lexical item enters the syntactic derivation, it passes through this computational component. Within this framework, the mechanism proposed for the state may also be applicable to other syntactically conditioned inflectional phenomena, such as agreement and case marking. We therefore suggest that similar computational models could be developed for these phenomena.

Certainly, there is more that can be detailed (e.g., the negative modality may also contribute to the state in some languages); we think that we are just touching upon the subject, but for the sake of conciseness, we will not elaborate further on this point. We are putting our finger on the fundamental concepts needed to explain how nouns can differ.

# 4 Conclusion

This article has reconsidered the state and proposed in its place a unified, computationally grounded account of the phenomenon traditionally confined to descriptions of the Semitic construct/absolute state and its Berber counterpart, the annexation/free state. Taking Riffian as its empirical anchor, the study has shown that surface variation in nominal marking, whether realised inter-word or in-situ, as illustrated by the cross-linguistic data, can be reduced to a single underlying mechanism: a syntactically conditioned shift between the unmarked and marked templates of a lexical item.

Formally, the state has been defined as a set-valued function mapping a syntactic context and an unmarked grammatical template onto a state-marked template and this formalisation is integrated within the TBMC model. The resulting symbolic learning framework learns grammatical transformations and predicts state alternations by means of algebraic operations on finite sets. This situates the proposal within the symbolic, algebraic branch of computational learning theory rather than within statistical or logic-based paradigms.

A dedicated learner, including the gradient condition, was shown to infer, from a sample of Riffian free- and annexation-state pairs, both the operand set governing the template shift and the two conditions — membership and applicability — required to select the appropriate branch function. The validation phase demonstrates that the learner correctly generalises beyond observed contexts while rejecting structurally incompatible configurations, providing empirical support for the proposed computational architecture, confirming that the learner reconstructs both observed and structurally equivalent unseen state pairings. This indicates that rule selection and template prediction can be evaluated as independent, jointly verifiable components of the model.

Beyond morphology, the theoretical consequences extend to syntactic analysis. Treating definite articles as intrinsic components of noun phrases rather than members of an independent determiner category offers a unified explanation of in-situ, inter-word and cross-linguistic variation while simplifying the analysis of nominal constructions. More generally, the framework suggests that many phenomena traditionally attributed to semantics, case marking, or language-specific syntactic operations can instead be derived from a common system prioritising a morphology-driven approach governed by the state.

This formal treatment was accompanied by a broader reassessment of markedness, redefined here pairwise, relative to the word- and meaning-formation process (inflection, derivation, state, etc) linking two items, and explicitly dissociated from formedness, the independent question of whether a morpheme is phonologically overt or null. On this basis, intrinsic inflectional markers, namely gender, number, countability and, in certain languages, definiteness, were distinguished from extrinsic markers such as most determiners, with the definite article reclassified as belonging to the morphological rather than the determiner domain. Together with the notion of subsettability proposed to unify countability and definiteness at the semantic level, this reclassification offers a principled account of the cross-linguistic complementarity between collective and definite marking, and of the differential treatment of proper and common nouns, without recourse to assumptions about the inherent definiteness of proper nouns.

Taken together, these results support a reinterpretation of nominal morphology. The state's cross-linguistic reach was illustrated through worked examples in English, French, Greek, Hebrew, Tagalog, Wolof, Bantu, Basque, Maori, etc. This spread of data supports the article's central claim that the state is not a parochial feature of Afroasiatic grammar but a general property of synthetic morphosyntax, patterning alongside — while remaining formally distinct from — agreement and grammatical case. Rather than proposing isolated analyses for particular languages or noun classes, this study provides a general framework capable of accounting for a broad range of morphosyntactic patterns through a common formal computational mechanism.

These proposals are not intended to be exhaustive. The generalisation of the state to the full range of synthetic languages will require systematic, language-by-language adaptation. However, the established learner and predictor will facilitate this task, as they can be applied without architectural modification. Although the empirical discussion has focused on nominal morphology, the proposed framework is not restricted to this domain. Future research should investigate the applicability of the state model to other grammatical categories and extend the computational learner to additional language families to evaluate the typological scope of the theory. Such work will help determine whether the morphology-driven organisation proposed here represents a general principle underlying grammatical systems more broadly.

## Declarations

### Funding

The author declares that this research did not receive funding.

### Availability of data and materials

The author confirms that the data supporting the findings of this study are available within the article.

### Competing Interests

The author declares that he has no conflict of interest.